\documentclass{article}
\usepackage{ijcai20}

\usepackage{times}

\usepackage{soul}
\usepackage{url}
\usepackage[hidelinks]{hyperref}
\usepackage[utf8]{inputenc}
\usepackage[small]{caption}
\usepackage{graphicx}
\usepackage{amsmath}
\usepackage{booktabs}
\usepackage{xcolor}
\usepackage{amsfonts}

\usepackage{multirow}
\usepackage{algorithm}
\usepackage{algorithmicx}
\usepackage{algpseudocode}
\usepackage{indentfirst} 
\usepackage{amsmath}
\usepackage{verbatim}

\title{Novelty Detection via Non-Adversarial Generative Network}

\author{
	Chengwei Chen$^1$\and
	Wang Yuan$^1$\and
	Yuan Xie$^{1}$\footnote{Contact Author}\and
	Yanyun Qu$^{2}$\and
	Yiqing Tao$^{1}$\And
	Haichuan Song$^{1}$\And
	Lizhuang Ma$^1$\\
	\affiliations
	$^1$East China Normal University\\
	$^2$XMU\\
	\emails
	\{52184501028, 51184501076,10161900112\}@stu.ecnu.edu.cn,
	xieyuan8589@foxmail.com,
	yyqu@xmu.edu.cn,
	hcsong@sei.ecnu.edu.cn,
	lzma@sei.ecnu.edu.cn
}

\begin{document}

\maketitle

\begin{abstract}

One-class novelty detection is the process of determining if a query example differs from the training examples (the target class). Most of previous strategies attempt to learn the real characteristics of target sample by using generative adversarial networks (GANs) methods. However, the training process of GANs remains challenging, suffering from instability issues such as mode collapse and vanishing gradients. In this paper,  by adopting non-adversarial generative networks, a novel \emph{decoder-encoder} framework is proposed for novelty detection task, insteading of classical encoder-decoder style. Under the non-adversarial framework, both latent space and image reconstruction space are jointly optimized, leading to a more stable training process with super fast convergence and lower training losses. During inference, inspired by cycleGAN, we design a new testing scheme to conduct image reconstruction, which is the reverse way of training sequence. Experiments show that our model has the clear superiority over cutting-edge novelty detectors and achieves the state-of-the-art results on the datasets.

\end{abstract}

\section{Introduction}

The objective of one-class novelty detection task is to detect the samples drawn far away from the learned distribution of training samples. Different from other computer vision tasks, in one-class novelty detection, only one-class samples are regarded as target class in the training process. During inference, the trained model distinguishes out-of-distribution samples from in-class samples. A lot of real world applications are related to this task, such as abnormal detection \cite{ravanbakhsh2017abnormal}, defect detection \cite{bibl2016method} and image denoising \cite{jifara2019medical}. However, since one-class novelty detection is based on the assumption that any negative samples could not be collected in training dataset, it poses a great challenge to this problem in practice. 

\begin{figure}[t]
	\centering
	\includegraphics[width=0.7\linewidth]{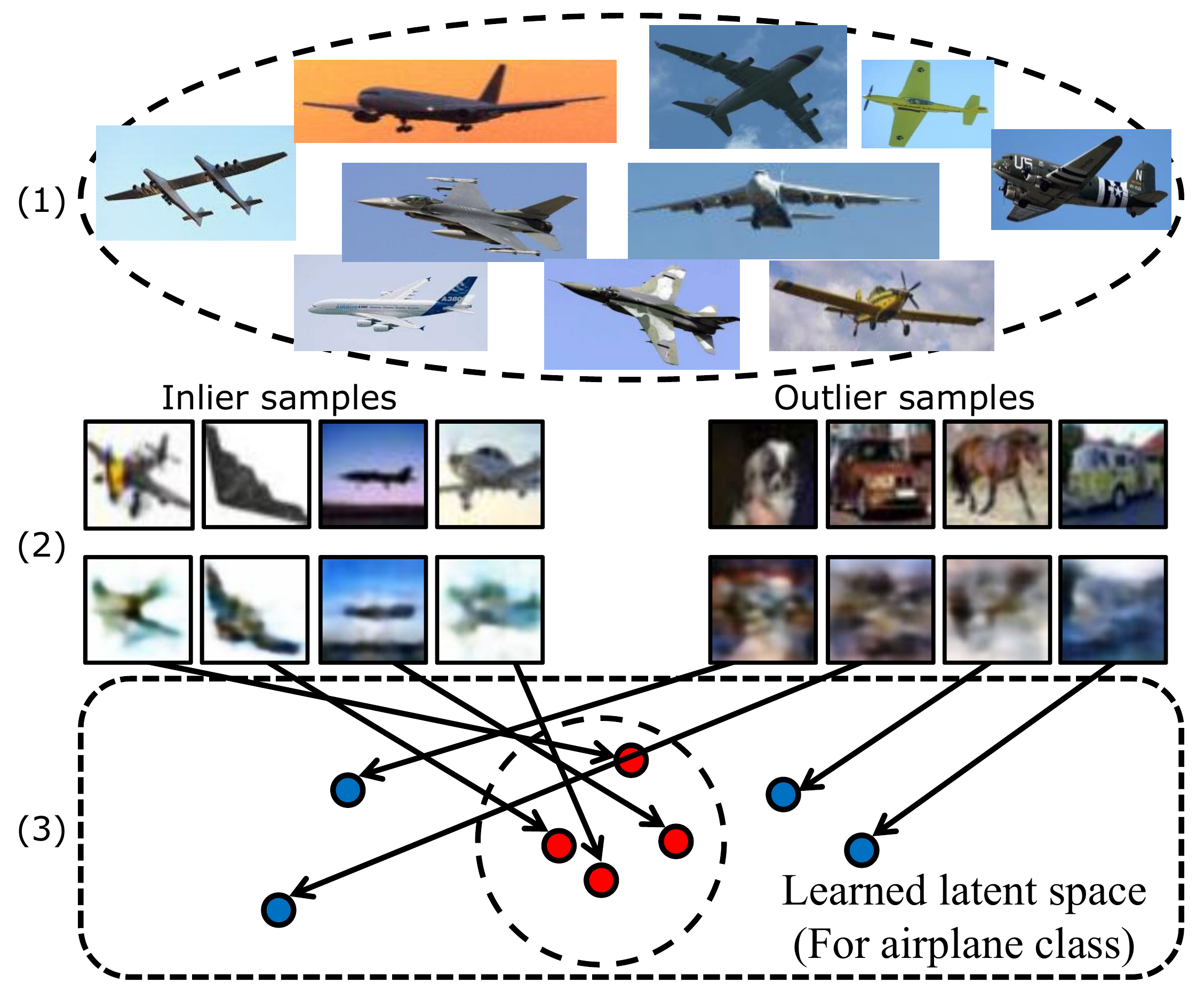}
	\caption{(1) shows the examples of airplane class. Only the “airplane” class samples is trained in the model. The first row of (2) presents test samples, and the second row includes the reconstructed images corresponding to the original input images. Since only the “airplane” class samples are trained in the model, it is more suitable for learned model to reconstruct inliers with high-quality. The outliers (dog, car, horse and truck) obtain reverse result.} 
	\label{vs}
\end{figure}

Due to the recent developments in generative adversarial networks (GANs) \cite{goodfellow2014generative}, several GAN-style methods \cite{sabokrou2018adversarially,eghbal2019mixture,zenati2018efficient} are proposed to detect novelty samples by using image reconstruction error. All of them consist of a generator (encoder-decoder) and a discriminator. Generator learns the real characteristics of target samples and generates the reconstructed image from learned latent space to fool the discriminator. Discriminator tries to distinguish the generated images from realistic input images. Both sub-networks compete with each other to achieve high-quality reconstructions that even the discriminator can not be determined. During the testing process, since the parameters of learned network is more suitable to reconstruct the normal samples, the out-of-distributions samples that naturally exhibit a higher reconstruction error, which is illustrated in Fig. \ref{vs}.

However, due to model collapse issues in GANs and multiple modes in the actual distribution of normal samples, GAN-style methods usually yield blurry reconstructions. The blurriness falsifies reconstruction errors, which is disastrous for reconstruction based novelty detection. More importantly, the imbalance of capability between generator and discriminator results in an unstable training process in any GAN-style frameworks  \cite{che2016mode}, degrading the performance to some extent.

Motivated by the above limitations, a novel deep network for one-class novelty detection has been proposed through non-adversarial learning strategy, which is based on the theory of generative latent optimization (GLO) \cite{bojanowski2017optimizing}. GLO framework merely provides us a deep convolutional decoder, which exploits the correspondence between each learned noise vector and the image sample that it represents by using simple reconstruction losses. To accommodate the novelty detection task, different from encoder-decoder style, we design a novel \emph{decoder-encoder} framework in a reverse manner, see Fig. \ref{architectures} for overview. Moreover, to generate discriminative latent representation and obtain high-quality reconstruction image, latent representation loss and image reconstruction loss are proposed to jointly optimize the latent feature space and image reconstruction space. To our best knowledge, we are the first to build a discriminative and robust latent representations in the novelty detection by \emph{non-adversarial generative networks}.

It is noteworthy that, in the training process, the proposed decoder-encoder structure is more simplified compared with the traditional GAN-based methods, leading to a super fast convergence for training that will be observed in Section 4.2. In contrast, GAN-style methods need careful choice of hyper-parameters and often multiple initializations. For testing process, inspired by cycleGAN \cite{zhu2017unpaired}, we design a new testing strategy to obtain reconstruction image, which is different from conventional encoder-decoder framework. Test sample is passed from the middle of the framework to encoder, and then the encoded latent feature is upscaled to reconstruct the image by decoder, which is the reverse of training sequence (decoder-encoder). Finally, we summarize the major contributions of this paper:

\begin{itemize}
	
	\item To avoid training issues of GANs, we propose a novel decoder-encoder style framework for novelty detection task by finding a meaningful organization of the noise vectors, from which target image could be well represented, all of this without the adversarial optimization scheme.
	
	\item To generate discriminative latent representations and obtain high-quality image reconstruction, latent representation loss and image reconstruction loss are proposed.  This framework can be easily trained with jointly multiple loss functions and converges quickly, leading to a more stable training process.
	
	\item We design a new testing approach to obtain reconstruction image for our decoder-encoder framework. Test sample is entered from the middle of framework to encoder, and then decoder reconstruct image from encoded latent feature which is the reverse of training process.
	
	\item We conduct the extensive evaluation of our method on several challenging datasets, where the experimental results demonstrate that our method outperforms many state-of-the-art competitors. 
	
\end{itemize}

\section{Related Work}

We give a brief review of novelty detection ranging from traditional methods to ones based on deep learning.

\begin{figure}[h]
	\centering
	\includegraphics[width=\linewidth]{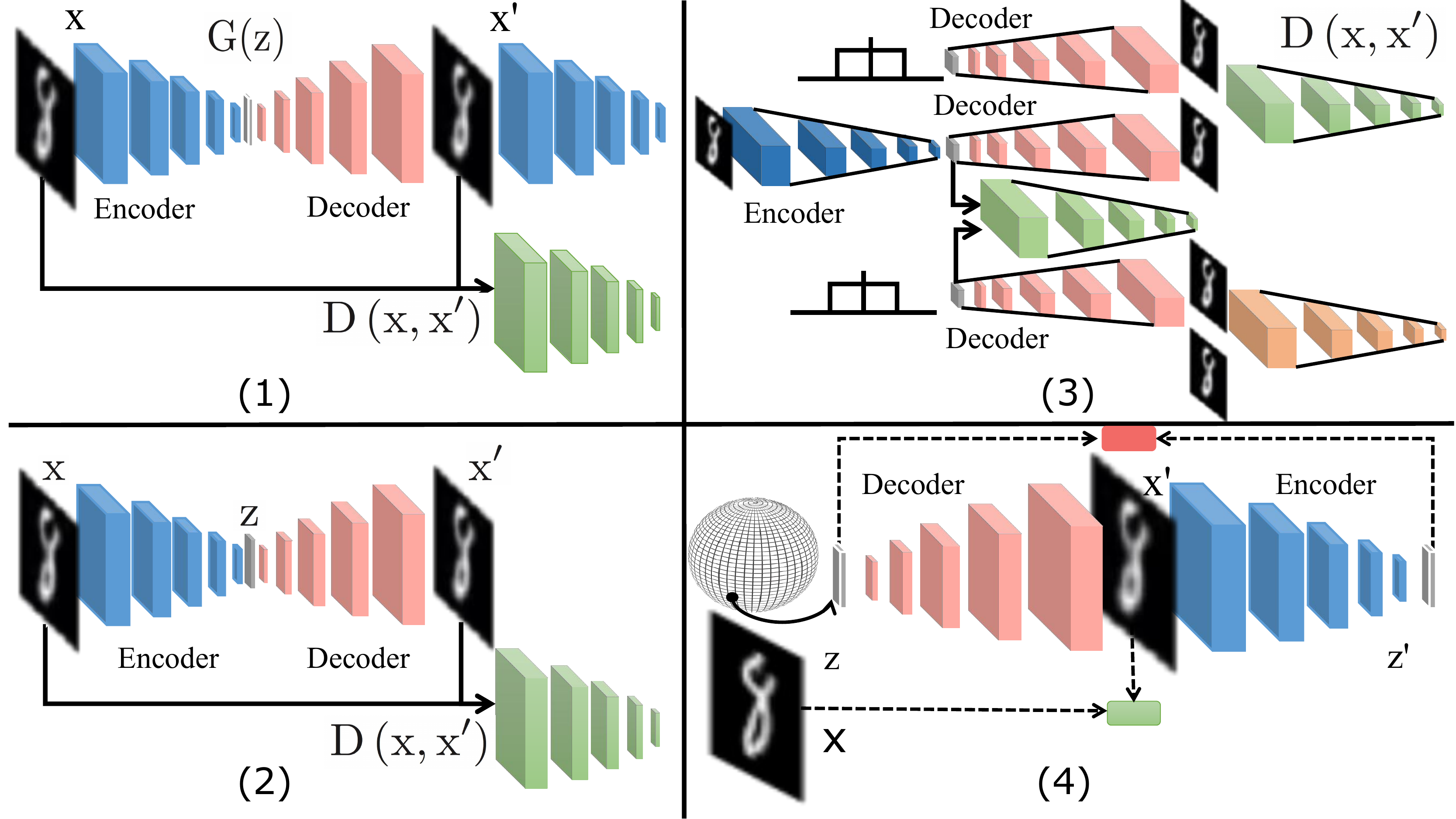}
	\caption{Comparison of the four models 1) GANomaly, 2) ALOCC, 3) OCGAN, 4) Our approach (decoder-encoder) }
	\label{architectures}
\end{figure}

\textbf{Self-representation: } Some previous researches, \emph{e.g.},  \cite{sabokrou2016video}, have presented that self-representation is an useful tool for novelty detection task or abnormal detection task. For example, due to the power of sparse representation and dictionary learning approach, researchers \cite{cong2013abnormal} used sparse representation to learn the dictionary of normal behaviors. In the process of testing, the patterns which have large reconstruction errors are considered as anomalous behaviors. In \cite{liu2010robust}, insteading of sparse representation, the authors proposed to employ a low-rank self-representation matrix in novelty detection task, which is penalized by the sum of unsquared self-representation errors, leading to a more robust detector.

\textbf{Autoencoders approaches:} 
Deep autoencoder \cite{vincent2010stacked,marchi2017deep,marchi2015novel} plays an essential role in novelty detection task. Autoencoder is trained by minimizing the distance between original image and generated image, so as to learn the real concept from target class. Since the auoencoder is trained with normal samples only, the parameters of model are not suitable to generate the abnormal samples, leading to a higher reconstruction error. However, the main objective of autoencoder is dimensionality reduction \cite{hinton2006reducing}, which is not the original intension for novelty detection. Besides, the biggest challenge of autoencoder is how to choose the right degree of compression. Hence, the ``compactness'' of latent representation relies on the hyperparameter of the model, which is difficult to determine.

\begin{figure*}[!h]
	\centering
	\includegraphics[width=0.85\linewidth]{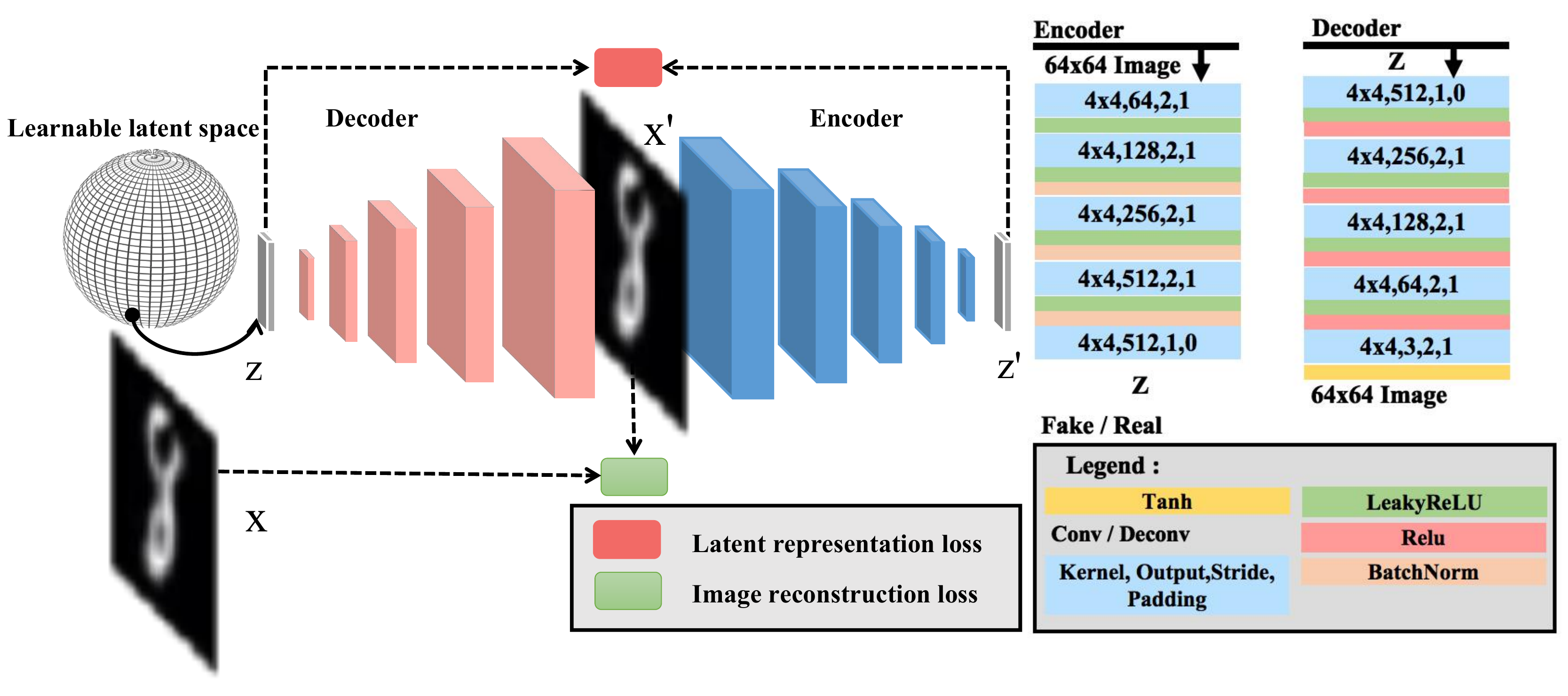}
	\caption{Our non-adversarial decoder-encoder framework consists of learnable latent space, a decoder subnetwork and an encoder subnetwork.}
	\label{architecture}
\end{figure*}

\textbf{Generative Adversarial Networks:} Apart from autoencoders, most of novelty detection works focus on GANs \cite{sabokrou2018adversarially,akcay2018ganomaly,perera2019ocgan}, as it shows in Fig. \ref{architectures} (1)-(3). The generator and discriminator are trained by competing with each other while collaborating to understand the underlying concept in the normal class. During testing process, the novelty samples are expected to obtain a higher image reconstruction error. However, the weakness of GAN style methods could be concluded as followed: 1) Mode dropping or mode collapse are the inherent issue in the GANs. GAN style methods ignore some part of the target distribution, leading to obtain blurry reconstructions. 2) The training process of GAN style methods with generator and discriminator is always unstable. In addition, it is hard to simultaneously train the multiple loss functions in the GAN methods.

\section{Proposed Method}

\subsection{Network Architecture}

The proposed non-adversarial generative network, which is shown in Fig. \ref{architecture}, consists of three components: learnable latent space, a decoder subnetwork and an encoder subnetwork. Firstly, the latent space is initialized from the PCA projections of training dataset. The decoder upscales the noise latent feature vector sampled from latent space to reconstruct the image, then the encoder tackles the problem by learning a mapping from generated image to a low dimensional representation. In order to map one learnable noise vector to each of the images in dataset, the distance between each image and corresponding generated image should be minimized. In addition, we also consider to optimize the learning representations in latent feature space by minimizing the distance between the random vector and the encoded latent feature of generated image from encoder.

\textbf{Learnable latent space: }Most of GANs methods \cite{dilokthanakul2016deep} choose a Normal distribution as the prior distribution to obtain the initialization of latent space. The random vector $z$ with $k$-dimension is initialized from normal distribution. All of these GANs methods detect novelty samples based on the assumption that  little or no multicollinearity exsits between the latent features. However, in practical, there are strong positive or negative correlations between each of latent features in the most of dataset. The performance of the model is always suffer from the multi-collinearity issue. 
Although many strategies have been proposed to tackle with this problem, principal component analysis (PCA) is a simple tool and widely used. Therefore, PCA is adopted to the initialization of random vector by reducing the dimensionality of data and exploiting new variables that are linear functions of those in the original data. In the initialization of representation space, a subset of training set is taken to fit the PCA, then we initialize latent vectors from the PCA projections of the dataset. Besides, for simplicity, we employ the unit sphere instead of the normal distribution on the $S(\sqrt{d}, d, 2)$ sphere.

\textbf{Decoder: }The images $\{x_{1}, \ldots, x_{N}\}$ of training set is trained in the proposed framework. Firstly, we initialize a set of random vectors $\{z_{1}, \ldots, z_{N}\}$ from the unit sphere described above, where each vector has $k$ dimensions. Secondly, we pair the training set with the each random vector. For each pair, the random vector is regard as a bottleneck feature in the
 conventional autoencoder. The main operator of the decoder $G_{d}(\cdot)$ is to upscale this random vector $z$ to obtain reconstruction image $x^{\prime}$.

\textbf{Encoder: }The encoder sub-network gets the generated image $x^{\prime}$ and passes it through the encoder $G_{e}(\cdot)$, which downscales $x^{\prime}$ by compressing it to another latent representation $z^{\prime}$ by convolutional layers followed by Batch Normalization (BN) and Leaky Relu activation, see the right part of Fig. \ref{architecture}.

\subsection{Overall Loss Function}

To train our model, we define a loss function in Eqn. (\ref{loss-function}) including two components, $i.e.$, the image reconstruction loss $\mathcal{L}_{irec}$ and latent representation loss $\mathcal{L}_{zrec}: $

\begin{equation}\label{loss-function}
	\mathcal{L}=w_{i} \mathcal{L}_{irec} + w_{z} \mathcal{L}_{zrec}
\end{equation}
where $w_{i}$ and $w_{z}$ are the weighting parameters balancing the impact of individual term to the overall object function. As for the first term, high-quality image reconstruction is obtained in the training process by minimizing the distance between original input image and generated image. To optimize latent representations, the distance between random latent vector $z$ and encoded latent vector $z^{\prime}$ from auxiliary encoder is minimized by latent representation loss $\mathcal{L}_{zrec}$.

\textbf{Image reconstruction loss: } In the optimization process of reconstruction space, a common choice of obtaining image reconstruction loss is MSE. However, using of MSE always yields blurry image. The Laplacian pyramid loss \cite{ling2006diffusion} is proposed to overcomes some of the known issues of the blurry images in GLO method. Therefore, we use the same reconstruction loss function to penalize the generator as follows,
\begin{equation}
\label{eq:3}
\mathcal{L}_{irec}=\sum_{j} 2^{2 j}\left|L^{j}(x)-L^{j}\left(x^{\prime}\right)\right|_{1},
\end{equation}
where $L^{j}(x)$ is the $j$-th level of the Laplacian pyramid representation of $x$.

\textbf{Latent representation loss: } Only for target class samples, latent representation loss will help the encoder to reconstruct the latent representation $z$ well from generated image $x^{\prime}$. Besides, the initialization of latent space might incur the distribution distortion in real latent feature space of training samples, the feature representation $z^{\prime}$ can be regarded as the anchor to prevent $z$ from drifting. Hence, we consider to minimize the distance between the random vector $z$ from the unit sphere and the encoded latent feature $z^{\prime}$ of generated image from encoder $G_{e}(x^{\prime})$ as follows.

\begin{equation}
\label{eq:4}
\mathcal{L}_{zrec}=\left\|z-G_{e}(x^{\prime})\right\|_{2}
\end{equation}

\subsection{Optimization: } 

\begin{figure}[h]
	\centering
	\includegraphics[width=\linewidth]{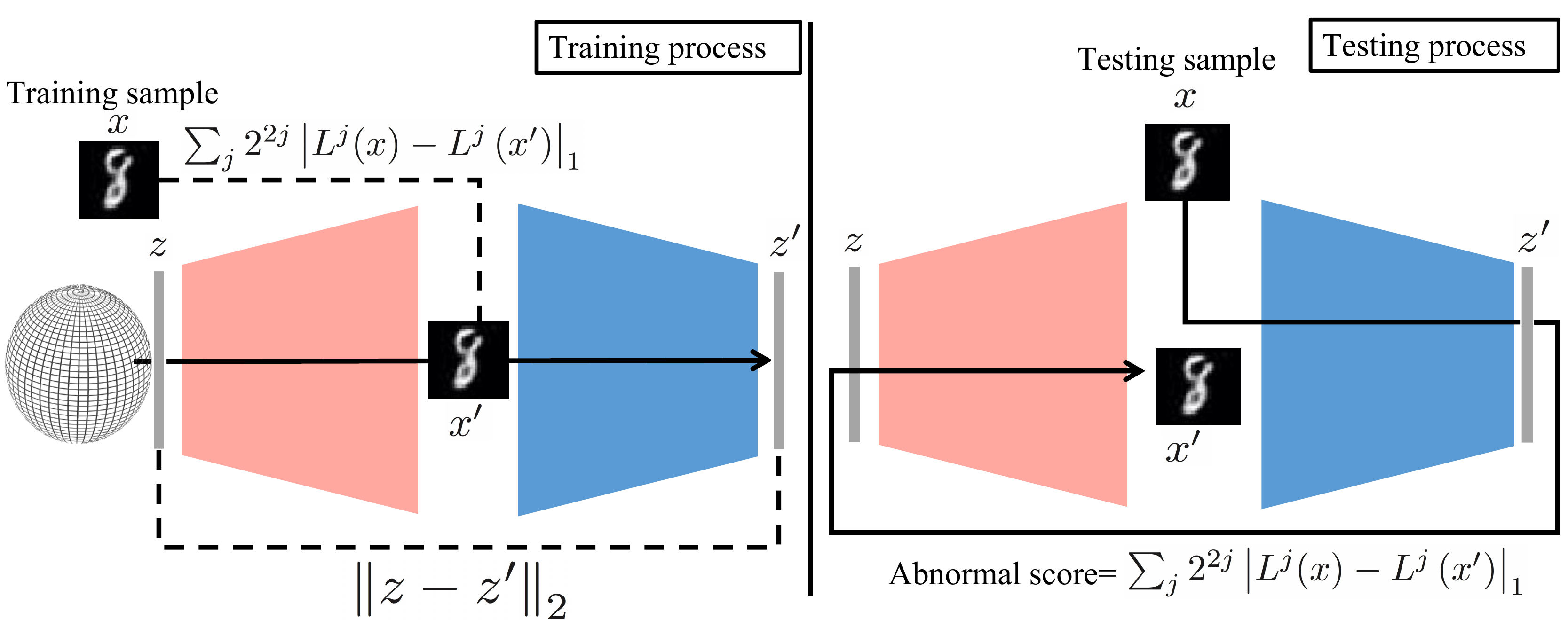}
	\caption{Left: To obtain the correspondence between each learned noise vector and the image sample that it represents, proposed method is trained by image reconstruction loss and latent representation loss. Right: In the testing process, to get the abnormal score for each sample, the test sample is passed from encoder to decoder, which is the reverse of the training process.}
	\label{process}
\end{figure}

We adopt our encoder and decoder subnetwork structure based on the DCGAN \cite{radford2015unsupervised}. In the process of optimization, the Stochastic Gradient Descent (SGD) is employed to optimize the parameters $\theta$ in the network. The learning rate of network is set to 0.002. As illustrated in Fig. \ref{process} (left), in the beginning of training, we initialize the random feature vector of proposed method by using the top 512 principal components of training set. We optimize the image reconstruction space by using image reconstruction error, which is calculated in laplacian pyramid loss function. $\ell_{2}$ loss is employed to optimize the latent space for all experiments.
During inference, as shown in Fig. \ref{process} (right), the test sample $x$ becomes the input of encoder and the decoder upscales the this feature vector to reconstruct the image from learned latent feature space. Finally, the abnormal score is calculated by image reconstruction error between testing image sample $x$ and corresponding generated image $x^{\prime}$. If image reconstruction error is larger than $T$, a predefined threshold, the test sample will be considered as a novelty instance.

\section{Experiments}

\subsection{Experimental Setting}

\begin{figure}[h]
	\centering
	\includegraphics[width=\linewidth]{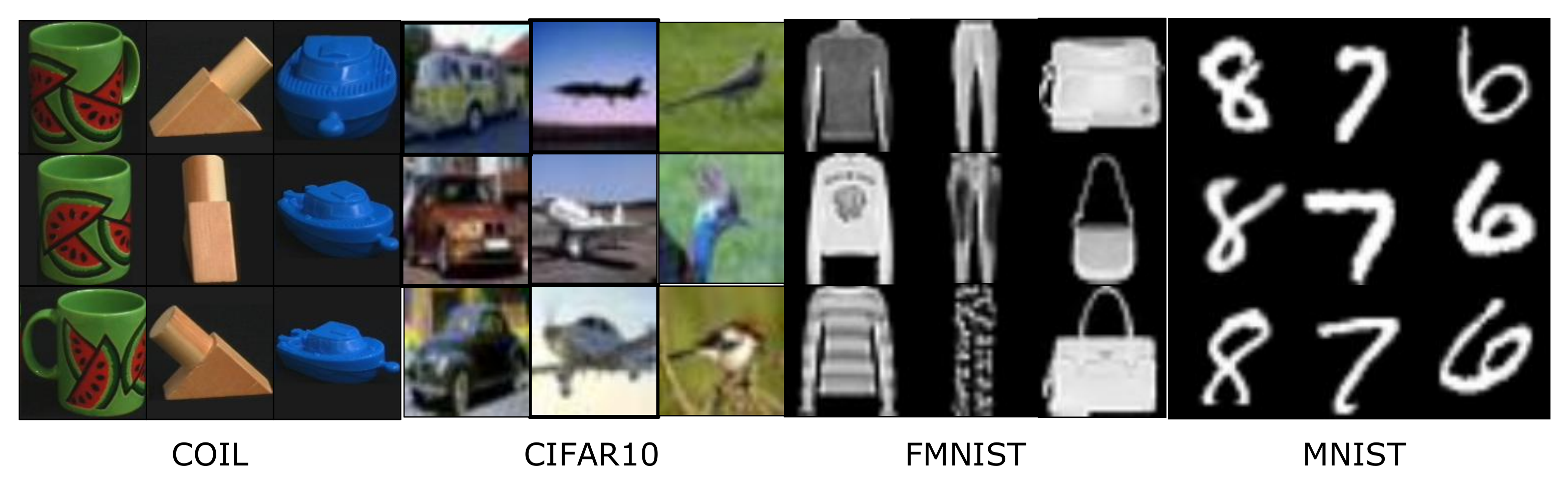}
	\caption{Representative images from four datasets. Each column presents the same class.}
	\label{dataset}
\end{figure}

\textbf{Datasets: }
COIL100 includes 100 objects with multiple different poses. Each class has less than one hundred images. MNIST dataset includes 60,000 handwritten digits from number 0 to number 9. The complexity of MNIST dataset is more challenging than COIL100. fMNIST dataset, consisting of 28x28 images of fashion apparels/accessories. CIFAR10 dataset also has ten classes with diverse content, background and complexity. The exmaples of these datasets are presented in Fig. \ref{dataset}.

In DCASE dataset, all of abnormal event audios are artificially mixed with background audios respectively ({\it i.e.,}home, bus, and train). 

\textbf{Evaluation Methodology: }Two protocols in the literature are proposed for one-class novelty detection \cite{perera2019ocgan}. 

Protocol 1 : The 80\% of in-class samples are regarded as normal class. The rest of 20\% of in-class samples is adopt in testing process. Out-of-class samples are serviced as abnormal class, which are randomly selected from testing dataset, constituting half of the test set.

Protocol 2 :  All of in-class samples from the training part of dataset is only used to train in the proposed method. Testing data of all classes are used for testing.

\textbf{Evaluation Measures:} The performance metrics we employed are Area Under Curve (AUC).

\subsection{Ablation Study}

\begin{table}[h]
	\scriptsize
	\centering 
	\caption{The effect of different initialization of latent space is evaluated in two dataset experiments.}
	\label{pcarandom}
	\begin{tabular}{c|c|c}
		\hline 
		&\textbf{CIFAR10} &\textbf{fMNIST}\\
		\hline 
		\textbf{PCA projection initialization}&\textbf{0.750}&\textbf{0.995}\\

		\textbf{Normal distribution	initialization} &0.667&0.990\\
		\hline 
	\end{tabular}	
\end{table}

In this section, since the initialization of representation space is key part of proposed framework in the process of training, it is necessary for us to evaluate different types of initialization strategies. Most of GAN-style methods consider to build a representation space from Normal distribution. In proposed non-adversarial generative networks, we initialize the latent space by fitting a subset of train set into PCA. In Tab. \ref{pcarandom}, we conduct the experiments by using PCA and Normal distribution  in protocol 2. It is obvious that when random latent vector is sampled from the PCA projections, performance of the proposed model is improved marginally by 8.3\% in CIFAR10 and 0.5\% in fMNIST respectively.

\begin{figure}[!h]
	\centering
	\includegraphics[width= \linewidth]{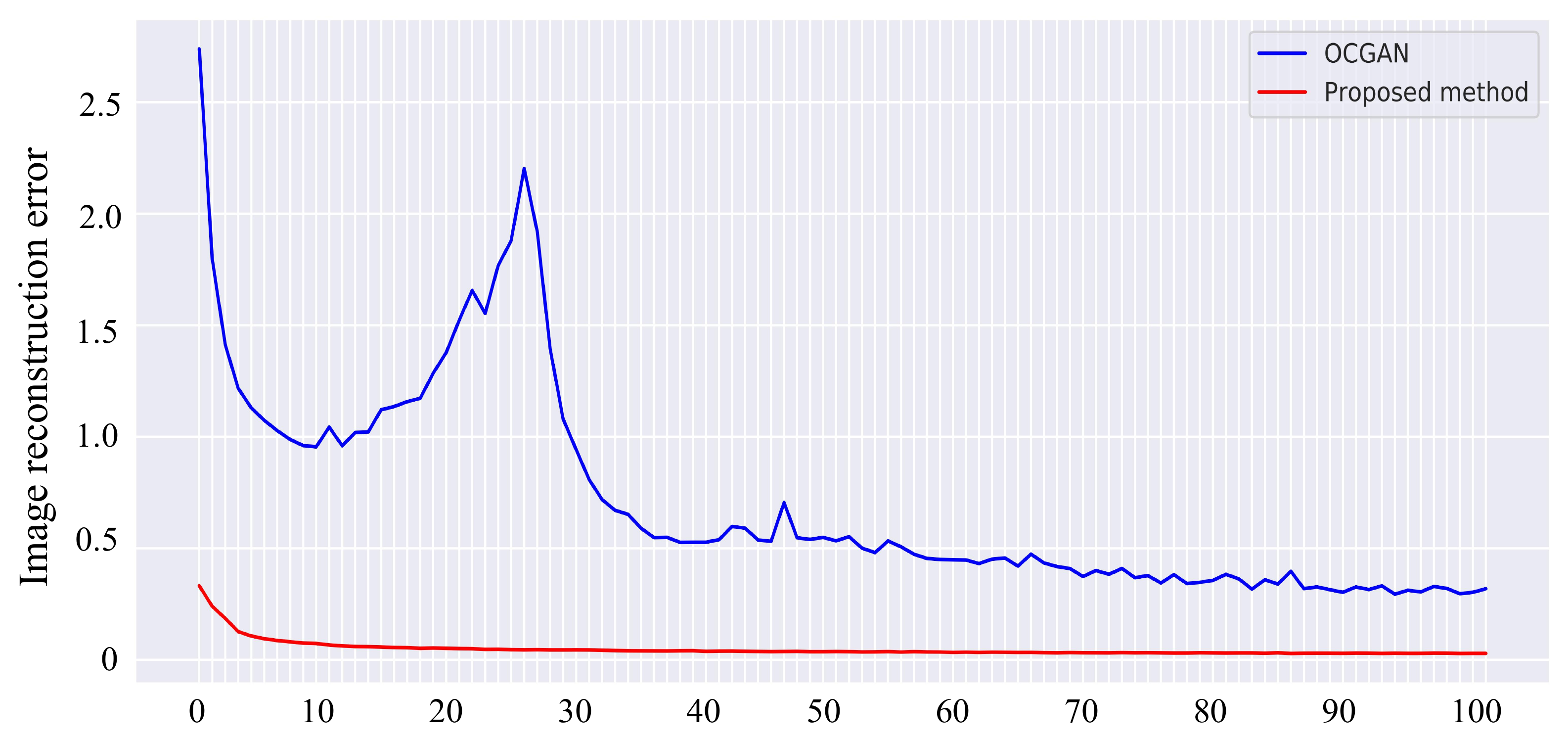}
	\caption{The visualization of training loss comparison between OCGAN and proposed method.}
	\label{imageloss}
\end{figure} 

\begin{figure}[!h]
	\centering
	\includegraphics[width=\linewidth]{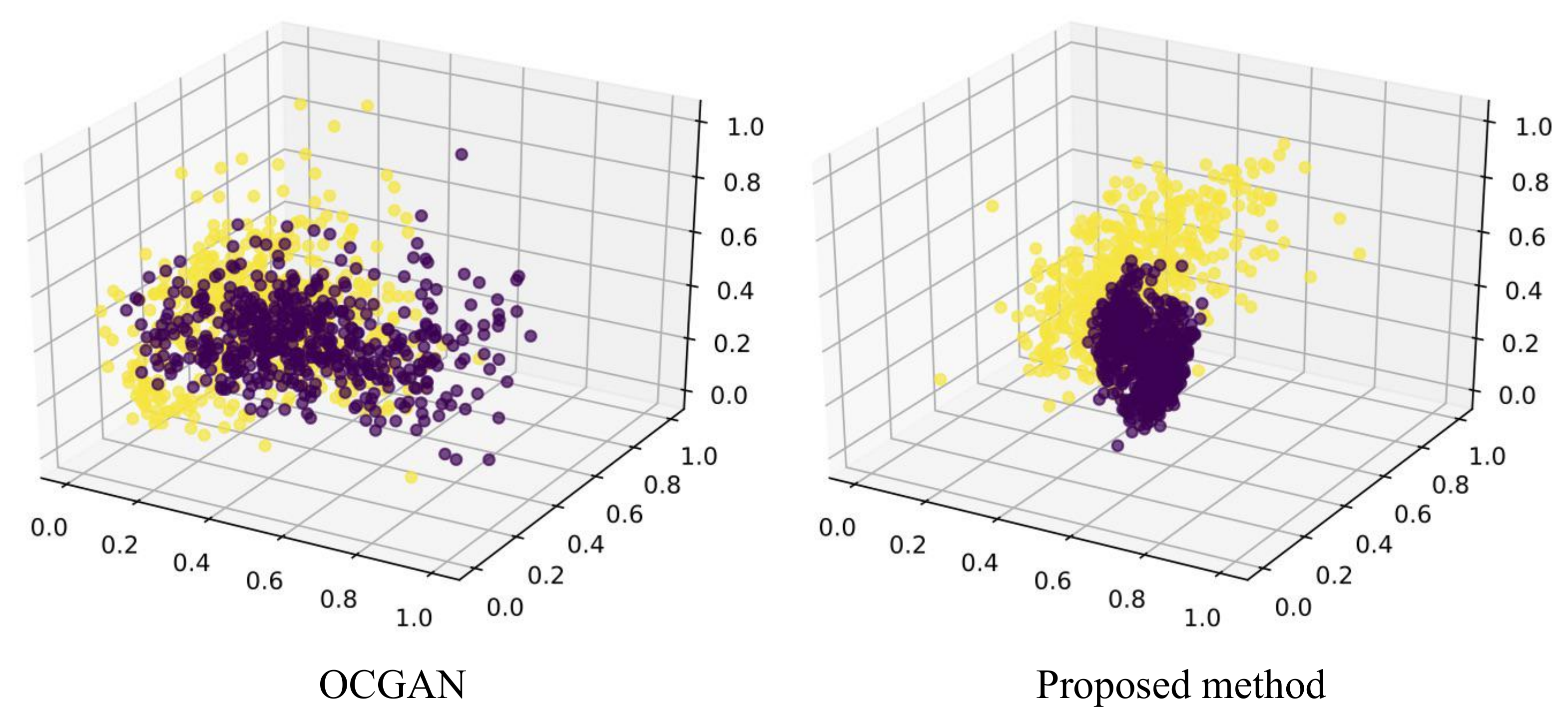}
	\caption{The visualization of latent space learned from target class (digit 8) by using OCGAN and proposed method.}
	\label{latent}
\end{figure}

To prove the stabilization effect of proposed method, we show the plot of training loss for OCGAN and proposed method, as illustrated in Fig. \ref{imageloss}. To sum up, our decoder-encoder style framework obtains a more stable training process and a faster convergence. To further present the optimization performance of latent space, as illustrated in Fig. \ref{latent}, where two learned latent space obtained from proposed method and OCGAN by training in MNIST (digit 8 regarded as normal class). We randomly select 200 in-class sample and 200 out-of-class samples from dataset for testing. The 3D scatter plot is used to visualize the latent representation $z$ for test samples, whose coordinates are calculated by applying the PCA for dimensional reduction. It is clear that the proposed latent feature of normal samples (purple dots) are more concentrated than that in the latent space of OCGAN. In addition, we can easily observe that the anomalous samples can not only be separated from normal samples but also can be mapped into the outside of margin of normal data in low-dimensional latent space in our method.

\subsection{Comparison with State-of-the-art Methods}

\begin{table}[]
	\scriptsize
	\centering 
	\caption{ Mean One-class novelty detection using Protocol 1.}
	\label{Protocol1}
	\begin{tabular}{c|c|c|c}
		\hline 
		&\textbf{MNIST} &\textbf{COIL}&\textbf{fMNIST}\\
\hline 
		\textbf{ALOCC DR ('18) }&0.88&0.809&0.753\\

		\textbf{ALOCC D ('18)}&0.82&0.686&0.601\\

		\textbf{DCAE ('14) }&0.899&0.949&0.908\\

		\textbf{GPND ('18)}&0.932&0.968&0.901 \\

		\textbf{OCGAN ('19)}&0.977&0.995&0.924 \\

		\textbf{Proposed method}&\textbf{0.985}&\textbf{1.0}&\textbf{0.995} \\
		\hline 
	\end{tabular}	
\end{table}
\textbf{Setup:} In this subsection, we consider to use both of protocols in the experiments. MNIST, COIL and fMNIST datasets are applied in the protocol 1. MNIST and CIFAR10 dataset are evaluated by the protocol 2. 

\textbf{Result:} When protocol 1 was used in MNIST dataset, proposed method yields an improvement of about 0.8\% compared to state-of-the-art method. In fMNIST and COIL dataset, the proposed method improves novelty detection performance by over 7.1\% and 0.5\%   compared to OCGAN  , using protocol 1. For experiments based on protocol 2, as shown in Table \ref{mnist}, for each digit, one of class is regarded as the normal class and the rest is served as outliers. We only use normal samples for training. Our method has not only registered a better average AUC value but also reported best AUC for individual classes in 7 out of 10 classes and a tie in 2 out of 10 classes.
\begin{figure}[h]
	\centering
	\includegraphics[width=\linewidth]{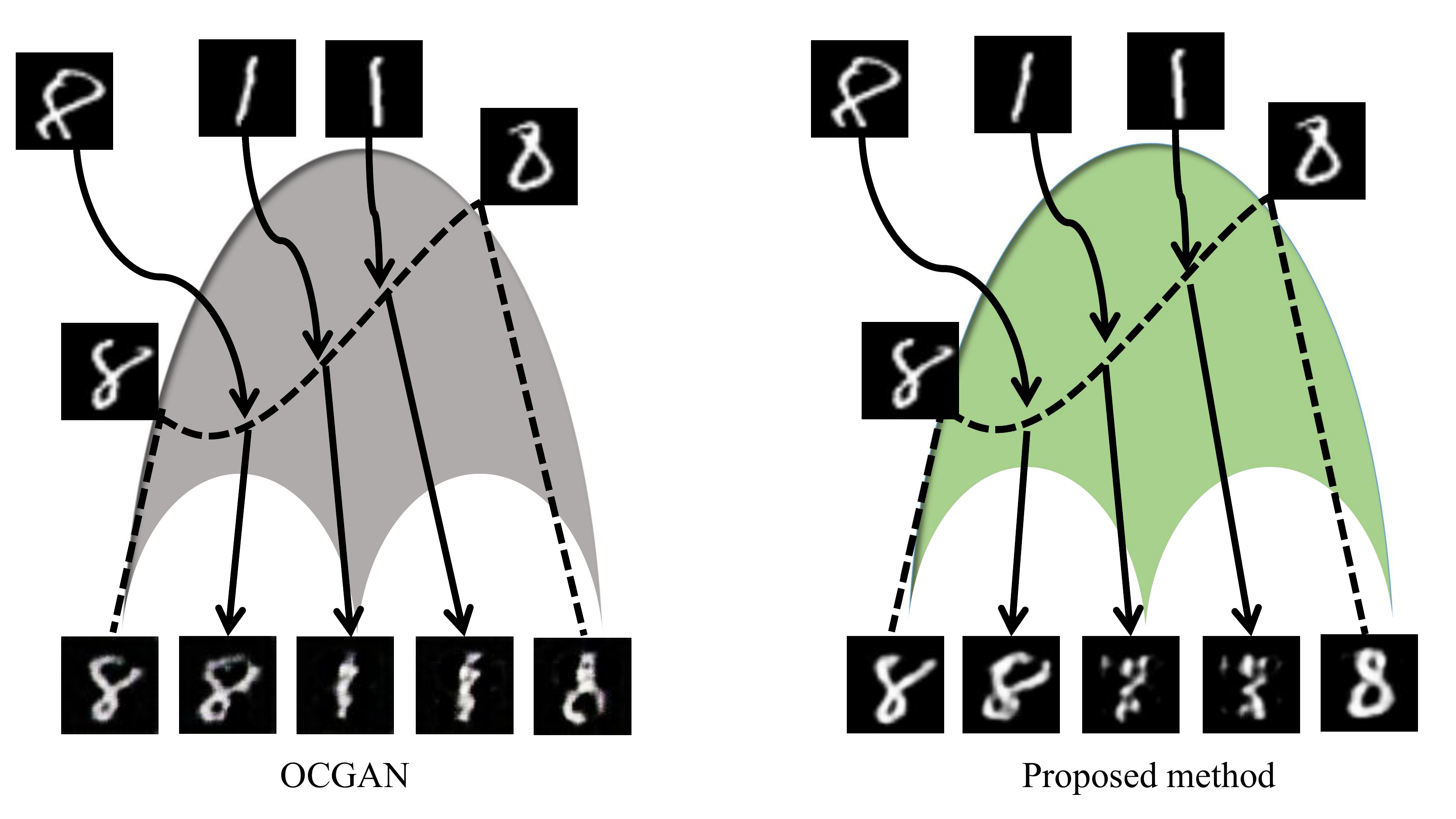}
	\caption{Due to the compactness of the proposed entire latent space corresponding to images from digit 8, all projections into the latent space in return produce images of digit 8, even for the out of distribution (digit 1) with higher reconstruction error.}
	\label{mnist_vs}
\end{figure} 

\begin{table*}[h]
	\scriptsize
	\centering 
	\renewcommand\tabcolsep{11.0pt} 
	\caption{One-class novelty detection results for MNIST dataset using Protocol 2.}
	\label{mnist}
	\begin{tabular}{c|c|c|c|c|c|c|c|c|c|c|c}
		\hline 
		&\textbf{0} &\textbf{1}&\textbf{2}&\textbf{3}&\textbf{4}&\textbf{5}&\textbf{6}&\textbf{7}&\textbf{8}&\textbf{9}&\textbf{MEAN}\\
		\hline 
		\textbf{OCSVM ('01)}&0.988&\textbf{0.999}&0.902&0.950&0.955&0.968&0.978&0.965&0.853&0.955&0.9513\\
		
		\textbf{KDE ('06)}&0.885&0.996&0.710&0.693&0.844&0.776&0.861&0.884&0.669&0.825&0.8143\\

		\textbf{DAE ('06)}&0.894&\textbf{0.999}&0.792&0.851&0.888&0.819&0.944&0.922&0.740&0.917&0.8766\\

		\textbf{VAE ('13)}&0.997&\textbf{0.999}&0.936&0.959&0.973&0.964&0.993&0.976&0.923&0.976&0.9696\\

		\textbf{Pix CNN ('16)}&0.531&0.995&0.476&0.517&0.739&0.542&0.592&0.789&0.340&0.662&0.6183\\

		\textbf{GAN ('17)}&0.926&0.995&0.805&0.818&0.823&0.803&0.890&0.898&0.817&0.887&0.8662\\

		\textbf{AND ('19)}&0.984&0.995&0.947&0.952&0.960&0.971&0.991&0.970&0.922&0.979&0.9671\\

		\textbf{AnoGAN ('17)}&0.966&0.992&0.850&0.887&0.894&0.883&0.947&0.935&0.849&0.924&0.9127\\

		\textbf{DSVDD ('18)}&0.980&0.997&0.917&0.919&0.949&0.885&0.983&0.946&0.939&0.965&0.9480\\

		\textbf{OCGAN ('19)}&\textbf{0.998}&\textbf{0.999}&0.942&0.963&\textbf{0.975} &0.980&0.991&0.981&0.939&\textbf{0.981}&0.9750\\

		\textbf{Proposed method}&\textbf{0.998} &\textbf{0.999}	&\textbf{0.987}&\textbf{0.986}&0.965&\textbf{0.989}&\textbf{0.998}	&\textbf{0.992}&\textbf{0.970}&	0.979	&\textbf{0.9830}\\
		\hline 
	\end{tabular}	
\end{table*}

\begin{table*}[h]
	\scriptsize
	\centering 
	\renewcommand\tabcolsep{10.0pt} 
	\caption{One-class novelty detection results for CIFAR10 dataset using Protocol 2. Plane and Car classes are annotated as Airplane and Automobile in CIFAR10.}
	\label{tab:intra-dataset-loss}
	\begin{tabular}{c|c|c|c|c|c|c|c|c|c|c|c}
		\hline 
		&\textbf{PLANE} &\textbf{CAR}&\textbf{BIRD}&\textbf{CAT}&\textbf{DEER}&\textbf{DOG}&\textbf{FROG}&\textbf{HORSE}&\textbf{SHIP}&\textbf{TRUCK}&\textbf{MEAN}\\
		\hline 
		\textbf{OCSVM ('01)}&0.630 &0.440 &0.649 &0.487& 0.735& 0.500 &0.725& 0.533& 0.649& 0.508& 0.5856\\

		\textbf{KDE ('06)}&0.658&0.520&0.657&0.497&0.727&0.496&0.758&0.564&0.680&0.540&0.6097\\

		\textbf{DAE ('06)}&0.411&0.478&0.616&0.562&0.728&0.513&0.688&0.497&0.487&0.378&0.5358\\
	
		\textbf{VAE ('13)}&0.700&0.386&0.679&0.535&0.748&0.523&0.687&0.493&0.696&0.386&0.5833\\

		\textbf{Pix CNN ('16)}&0.788&0.428&0.617&0.574&0.511&0.571&0.422&0.454&0.715&0.426&0.5506\\
	
		\textbf{GAN ('17)}&0.708&0.458&0.664&0.510&0.722&0.505&0.707&0.471&0.713&0.458&0.5916\\

		\textbf{AND ('19)}&0.717&0.494&0.662&0.527&0.736&0.504&0.726&0.560&0.680&0.566&0.6172\\

		\textbf{AnoGAN ('17)}&0.671&0.547&0.529&0.545&0.651&0.603&0.585&0.625&0.758&0.665&0.6179\\

		\textbf{DSVDD ('18)}&0.617&\textbf{0.659}&0.508&0.591&0.609&\textbf{0.657}&0.677&\textbf{0.673}&0.759&\textbf{0.731}&0.6481\\

		\textbf{OCGAN ('19)}&0.757&0.531&0.640&0.620&0.723&0.620&0.723&0.575&0.820&0.554&0.6566 \\

		\textbf{Proposed method}&\textbf{0.962}	&0.638&\textbf{0.725}&\textbf{0.643}&\textbf{0.873}&	0.638&	\textbf{0.883}&	0.584	&\textbf{0.935}&	0.645&	\textbf{0.7501}\\
		\hline 
	\end{tabular}	
\end{table*}

Compared with OCGAN, proposed method could learn the most representive concept of target class in latent manifold by our decoder-encoder framework, making the normalities and anomalies more separable and obtaining a more accurate detector. To make an intuitive comparison, two learned latent manifolds of (digit 8) images are obtained by OCGAN and proposed method respectively, as shown in Fig. \ref{mnist_vs}. In our method, all projections into the latent manifold in return produce images of digit 8, even for the out of distribution samples (digit 1) with higher reconstruction error. The main reason is that our method could capture the real concept of target class (digit 8) in the entire latent manifold under the constraint,  leading to a more compact learned latent manifold, from which the abnormal samples could not be represented well. While, as for the latent space learned from OCGAN, the recovery of the digital 1 is more like to itself, which is harmful to distinguish.

In comparison, CIFAR10 is not an aligned dataset and it contains objects of the given class across very different settings. As a result, we obtain state-of-art results  with the proposed method where we recorded average AUC of 0.750.

\begin{figure}[h]
	\centering
	\includegraphics[width=\linewidth,height=0.3\linewidth]{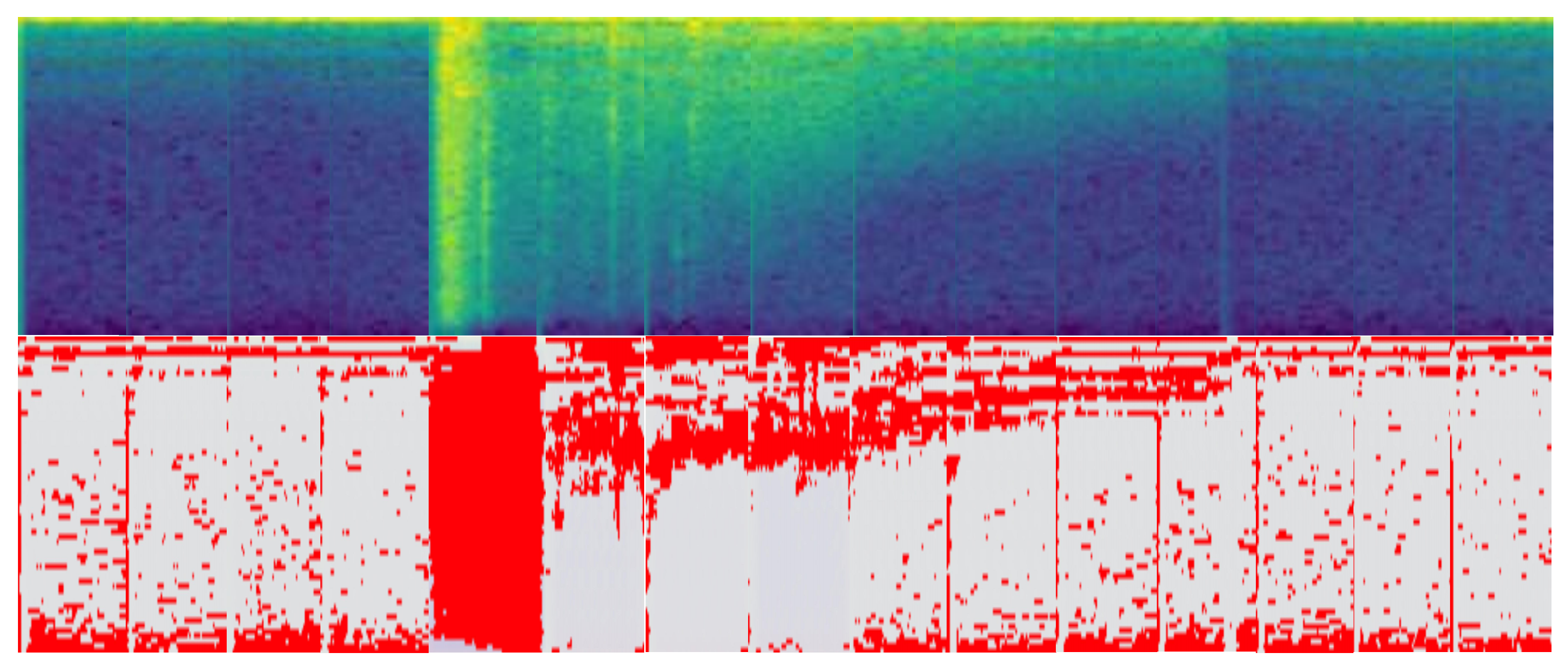}
	\caption{
		First row: the original spectrogram of acoustic signal. Second row: corresponding maps which display the differences between original spectrogram image (the first row) and corresponding generated images by red point.}
	\label{spc}
\end{figure}

For acoustic anomaly detection task, since anomalous sounds might indicate a rare and unexpected event, their prompt detection can possibly prevent such problems. Proposed method aims at distinguishing abnormal acoustic signals from the normal ones. We also evaluate  the DCASE dataset in acoustic novelty detection task by using protocol 2. DCASE dataset includes three different abnormal event ({\it i.e.,} gunshot, babycry and glassbreak). All of these abnormal event audios are artificially mixed with background audios respectively which includes 15 different kinds of environmental settings ({\it i.e.,}home, bus, and train). The performance of three models across the 15 datasets is shown in Table \ref{voice}. We find that the proposed model consistently outperforms the other models in almost all datasets, except  home scenarios. To further present the effectiveness of proposed method, in Fig. \ref{spc}, it shows the pixel-wise difference between original spectrogram images and reconstruction image in gunshot voice mixed with car background audios. Most of abnormal regions are detected by our method, which is labeled by red point.

\begin{table}[]
	\scriptsize
	\centering
	\caption{AUC scores for all methods on each dataset.}
	\label{voice}
	\begin{tabular}{c|c|c|c}
		\hline 
		\textbf{Dataset}&\textbf{CAE ('16)}&\textbf{WaveNet ('19)}&\textbf{Proposed method} \\
		\hline 
		\textbf{Beach}&0.69&0.72&\textbf{0.86}\\

		\textbf{Bus}&0.79&0.83&\textbf{0.96}\\

		\textbf{Cafe/restaurant}&0.69&0.76&\textbf{0.78}\\

		\textbf{Car}&0.79&0.82&\textbf{0.99}\\

		\textbf{City center}&0.75&0.82&\textbf{0.90}\\

		\textbf{Forest path}&0.65&0.72&\textbf{0.80}\\

		\textbf{Grocery store}&0.71&0.77&\textbf{0.95}\\

		\textbf{Home}&0.69&\textbf{0.69}&0.68\\

		\textbf{Library}&0.59&0.67&\textbf{0.97}\\

		\textbf{Metro station}&0.74&0.79&\textbf{0.93}\\

		\textbf{Office}&0.78&0.78&\textbf{0.94}\\

		\textbf{Park}&0.70&0.80&\textbf{0.99}\\

		\textbf{Residential area}&0.73&0.78&\textbf{0.81}\\

		\textbf{Train}&0.82&0.84&\textbf{0.95}\\

		\textbf{Tram}&0.80&0.87&\textbf{0.97}\\
		\hline 
	\end{tabular}
\end{table}

\section{Conclusion}

To avoid training issues of GANs and better represent the target image, a novel decoder-encoder style framework are proposed to find the meaningful organization of the noise vectors in a non-adversarial optimization manner. In the
 training process, to achieve discriminative latent representations and high-quality image reconstruction, latent representation loss and image reconstruction loss jointly regularize the reconstruction space and latent space, leading to a quick convergence and a more stable training process with lower losses. During testing process, a new testing strategy is adopted to obtain the reconstruction image, which is reverse of training process. Extensive experiments have been conducted on some datasets, showing effectiveness of model trained on our novel decoder-encoder framework and the benefit of using regularizers.

\clearpage
\bibliographystyle{named}
\bibliography{ijcai20}

\begin{thebibliography}{}

\bibitem[\protect\citeauthoryear{Akcay \bgroup \em et al.\egroup
  }{2018}]{akcay2018ganomaly}
Samet Akcay, Amir Atapour-Abarghouei, and Toby~P Breckon.
\newblock Ganomaly: Semi-supervised anomaly detection via adversarial training.
\newblock In {\em Asian Conference on Computer Vision}, pages 622--637.
  Springer, 2018.

\bibitem[\protect\citeauthoryear{Bibl \bgroup \em et al.\egroup
  }{2016}]{bibl2016method}
Andreas Bibl, Kapil~V Sakariya, Charles~R Griggs, and James~Michael Perkins.
\newblock Method of fabricating a light emitting diode display with integrated
  defect detection test, February~2 2016.
\newblock US Patent 9,252,375.

\bibitem[\protect\citeauthoryear{Bojanowski \bgroup \em et al.\egroup
  }{2017}]{bojanowski2017optimizing}
Piotr Bojanowski, Armand Joulin, David Lopez-Paz, and Arthur Szlam.
\newblock Optimizing the latent space of generative networks.
\newblock {\em arXiv preprint arXiv:1707.05776}, 2017.

\bibitem[\protect\citeauthoryear{Che \bgroup \em et al.\egroup
  }{2016}]{che2016mode}
Tong Che, Yanran Li, Athul~Paul Jacob, Yoshua Bengio, and Wenjie Li.
\newblock Mode regularized generative adversarial networks.
\newblock {\em arXiv preprint arXiv:1612.02136}, 2016.

\bibitem[\protect\citeauthoryear{Cong \bgroup \em et al.\egroup
  }{2013}]{cong2013abnormal}
Yang Cong, Junsong Yuan, and Ji~Liu.
\newblock Abnormal event detection in crowded scenes using sparse
  representation.
\newblock {\em Pattern Recognition}, 46(7):1851--1864, 2013.

\bibitem[\protect\citeauthoryear{Dilokthanakul \bgroup \em et al.\egroup
  }{2016}]{dilokthanakul2016deep}
Nat Dilokthanakul, Pedro~AM Mediano, Marta Garnelo, Matthew~CH Lee, Hugh
  Salimbeni, Kai Arulkumaran, and Murray Shanahan.
\newblock Deep unsupervised clustering with gaussian mixture variational
  autoencoders.
\newblock {\em arXiv preprint arXiv:1611.02648}, 2016.

\bibitem[\protect\citeauthoryear{Eghbal-zadeh \bgroup \em et al.\egroup
  }{2019}]{eghbal2019mixture}
Hamid Eghbal-zadeh, Werner Zellinger, and Gerhard Widmer.
\newblock Mixture density generative adversarial networks.
\newblock In {\em Proceedings of the IEEE Conference on Computer Vision and
  Pattern Recognition}, pages 5820--5829, 2019.

\bibitem[\protect\citeauthoryear{Goodfellow \bgroup \em et al.\egroup
  }{2014}]{goodfellow2014generative}
Ian Goodfellow, Jean Pouget-Abadie, Mehdi Mirza, Bing Xu, David Warde-Farley,
  Sherjil Ozair, Aaron Courville, and Yoshua Bengio.
\newblock Generative adversarial nets.
\newblock In {\em Advances in neural information processing systems}, pages
  2672--2680, 2014.

\bibitem[\protect\citeauthoryear{Hinton and
  Salakhutdinov}{2006}]{hinton2006reducing}
Geoffrey~E Hinton and Ruslan~R Salakhutdinov.
\newblock Reducing the dimensionality of data with neural networks.
\newblock {\em science}, 313(5786):504--507, 2006.

\bibitem[\protect\citeauthoryear{Jifara \bgroup \em et al.\egroup
  }{2019}]{jifara2019medical}
Worku Jifara, Feng Jiang, Seungmin Rho, Maowei Cheng, and Shaohui Liu.
\newblock Medical image denoising using convolutional neural network: a
  residual learning approach.
\newblock {\em The Journal of Supercomputing}, 75(2):704--718, 2019.

\bibitem[\protect\citeauthoryear{Ling and Okada}{2006}]{ling2006diffusion}
Haibin Ling and Kazunori Okada.
\newblock Diffusion distance for histogram comparison.
\newblock In {\em 2006 IEEE Computer Society Conference on Computer Vision and
  Pattern Recognition (CVPR'06)}, volume~1, pages 246--253. IEEE, 2006.

\bibitem[\protect\citeauthoryear{Liu \bgroup \em et al.\egroup
  }{2010}]{liu2010robust}
Guangcan Liu, Zhouchen Lin, and Yong Yu.
\newblock Robust subspace segmentation by low-rank representation.
\newblock In {\em ICML}, volume~1, page~8, 2010.

\bibitem[\protect\citeauthoryear{Marchi \bgroup \em et al.\egroup
  }{2015}]{marchi2015novel}
Erik Marchi, Fabio Vesperini, Florian Eyben, Stefano Squartini, and Bj{\"o}rn
  Schuller.
\newblock A novel approach for automatic acoustic novelty detection using a
  denoising autoencoder with bidirectional lstm neural networks.
\newblock In {\em 2015 IEEE International Conference on Acoustics, Speech and
  Signal Processing (ICASSP)}, pages 1996--2000. IEEE, 2015.

\bibitem[\protect\citeauthoryear{Marchi \bgroup \em et al.\egroup
  }{2017}]{marchi2017deep}
Erik Marchi, Fabio Vesperini, Stefano Squartini, and Bj{\"o}rn Schuller.
\newblock Deep recurrent neural network-based autoencoders for acoustic novelty
  detection.
\newblock {\em Computational intelligence and neuroscience}, 2017, 2017.

\bibitem[\protect\citeauthoryear{Perera \bgroup \em et al.\egroup
  }{2019}]{perera2019ocgan}
Pramuditha Perera, Ramesh Nallapati, and Bing Xiang.
\newblock Ocgan: One-class novelty detection using gans with constrained latent
  representations.
\newblock In {\em Proceedings of the IEEE Conference on Computer Vision and
  Pattern Recognition}, pages 2898--2906, 2019.

\bibitem[\protect\citeauthoryear{Radford \bgroup \em et al.\egroup
  }{2015}]{radford2015unsupervised}
Alec Radford, Luke Metz, and Soumith Chintala.
\newblock Unsupervised representation learning with deep convolutional
  generative adversarial networks.
\newblock {\em arXiv preprint arXiv:1511.06434}, 2015.

\bibitem[\protect\citeauthoryear{Ravanbakhsh \bgroup \em et al.\egroup
  }{2017}]{ravanbakhsh2017abnormal}
Mahdyar Ravanbakhsh, Moin Nabi, Enver Sangineto, Lucio Marcenaro, Carlo
  Regazzoni, and Nicu Sebe.
\newblock Abnormal event detection in videos using generative adversarial nets.
\newblock In {\em 2017 IEEE International Conference on Image Processing
  (ICIP)}, pages 1577--1581. IEEE, 2017.

\bibitem[\protect\citeauthoryear{Sabokrou \bgroup \em et al.\egroup
  }{2016}]{sabokrou2016video}
Mohammad Sabokrou, Mahmood Fathy, and Mojtaba Hoseini.
\newblock Video anomaly detection and localisation based on the sparsity and
  reconstruction error of auto-encoder.
\newblock {\em Electronics Letters}, 52(13):1122--1124, 2016.

\bibitem[\protect\citeauthoryear{Sabokrou \bgroup \em et al.\egroup
  }{2018}]{sabokrou2018adversarially}
Mohammad Sabokrou, Mohammad Khalooei, Mahmood Fathy, and Ehsan Adeli.
\newblock Adversarially learned one-class classifier for novelty detection.
\newblock In {\em Proceedings of the IEEE Conference on Computer Vision and
  Pattern Recognition}, pages 3379--3388, 2018.

\bibitem[\protect\citeauthoryear{Vincent \bgroup \em et al.\egroup
  }{2010}]{vincent2010stacked}
Pascal Vincent, Hugo Larochelle, Isabelle Lajoie, Yoshua Bengio, and
  Pierre-Antoine Manzagol.
\newblock Stacked denoising autoencoders: Learning useful representations in a
  deep network with a local denoising criterion.
\newblock {\em Journal of machine learning research}, 11(Dec):3371--3408, 2010.

\bibitem[\protect\citeauthoryear{Zenati \bgroup \em et al.\egroup
  }{2018}]{zenati2018efficient}
Houssam Zenati, Chuan~Sheng Foo, Bruno Lecouat, Gaurav Manek, and
  Vijay~Ramaseshan Chandrasekhar.
\newblock Efficient gan-based anomaly detection.
\newblock {\em arXiv preprint arXiv:1802.06222}, 2018.

\bibitem[\protect\citeauthoryear{Zhu \bgroup \em et al.\egroup
  }{2017}]{zhu2017unpaired}
Jun-Yan Zhu, Taesung Park, Phillip Isola, and Alexei~A Efros.
\newblock Unpaired image-to-image translation using cycle-consistent
  adversarial networks.
\newblock In {\em Proceedings of the IEEE international conference on computer
  vision}, pages 2223--2232, 2017.

\end{thebibliography}


\begin{thebibliography}{}

\bibitem[\protect\citeauthoryear{Abelson \bgroup \em et al.\egroup
  }{1985}]{abelson-et-al:scheme}
Harold Abelson, Gerald~Jay Sussman, and Julie Sussman.
\newblock {\em Structure and Interpretation of Computer Programs}.
\newblock MIT Press, Cambridge, Massachusetts, 1985.

\bibitem[\protect\citeauthoryear{Baumgartner \bgroup \em et al.\egroup
  }{2001}]{bgf:Lixto}
Robert Baumgartner, Georg Gottlob, and Sergio Flesca.
\newblock Visual information extraction with {Lixto}.
\newblock In {\em Proceedings of the 27th International Conference on Very
  Large Databases}, pages 119--128, Rome, Italy, September 2001. Morgan
  Kaufmann.

\bibitem[\protect\citeauthoryear{Brachman and
  Schmolze}{1985}]{brachman-schmolze:kl-one}
Ronald~J. Brachman and James~G. Schmolze.
\newblock An overview of the {KL-ONE} knowledge representation system.
\newblock {\em Cognitive Science}, 9(2):171--216, April--June 1985.

\bibitem[\protect\citeauthoryear{Gottlob \bgroup \em et al.\egroup
  }{2002}]{gls:hypertrees}
Georg Gottlob, Nicola Leone, and Francesco Scarcello.
\newblock Hypertree decompositions and tractable queries.
\newblock {\em Journal of Computer and System Sciences}, 64(3):579--627, May
  2002.

\bibitem[\protect\citeauthoryear{Gottlob}{1992}]{gottlob:nonmon}
Georg Gottlob.
\newblock Complexity results for nonmonotonic logics.
\newblock {\em Journal of Logic and Computation}, 2(3):397--425, June 1992.

\bibitem[\protect\citeauthoryear{Levesque}{1984a}]{levesque:functional-foundations}
Hector~J. Levesque.
\newblock Foundations of a functional approach to knowledge representation.
\newblock {\em Artificial Intelligence}, 23(2):155--212, July 1984.

\bibitem[\protect\citeauthoryear{Levesque}{1984b}]{levesque:belief}
Hector~J. Levesque.
\newblock A logic of implicit and explicit belief.
\newblock In {\em Proceedings of the Fourth National Conference on Artificial
  Intelligence}, pages 198--202, Austin, Texas, August 1984. American
  Association for Artificial Intelligence.

\bibitem[\protect\citeauthoryear{Nebel}{2000}]{nebel:jair-2000}
Bernhard Nebel.
\newblock On the compilability and expressive power of propositional planning
  formalisms.
\newblock {\em Journal of Artificial Intelligence Research}, 12:271--315, 2000.

\end{thebibliography}

\end{document}